\documentclass[journal,twoside,web]{ieeecolor}
\usepackage{jsen}
\usepackage{amsmath,amssymb,amsfonts}
\usepackage{algorithmic}
\usepackage{graphicx}
\usepackage{textcomp}
\usepackage{wrapfig}
\usepackage{amsmath}
\usepackage{subcaption}
\usepackage{booktabs}
\usepackage{multirow}
\usepackage{array}
\usepackage{tabularx}
\usepackage[pagebackref=false, breaklinks=true, colorlinks=true, citecolor=blue, bookmarks=false]{hyperref}

\def\BibTeX{{\rm B\kern-.05em{\sc i\kern-.025em b}\kern-.08em
    T\kern-.1667em\lower.7ex\hbox{E}\kern-.125emX}}
\definecolor{abstractbg}{rgb}{0.89804,0.94510,0.83137}
\begin{document}
\title{$O^{2}Former$: Direction-Aware and Multi-Scale Query Enhancement for SAR Ship Instance Segmentation}
\author{Fei Gao, Yunrui Li, Xiaoyu He, Jinping Sun, \IEEEmembership{Member, IEEE} and Jun Wang
\thanks{ }
\thanks{Fei Gao, Yunrui Li, Xiaoyu He, Jinping Sun and Jun wang are with the College of Electronic information engineering, Beihang University, Beijing 100083, China  (e-mail:  feigao2000@163.com; yunrui@buaa.edu.cn; hexiaoyu@buaa.edu.cn; sunjinping@buaa.edu.cn; wangj203@buaa.edu.cn).}
}
\IEEEtitleabstractindextext{%
\fcolorbox{abstractbg}{abstractbg}{%
\begin{minipage}{\textwidth}%
\begin{wrapfigure}[12]{r}{3.1in}%
\includegraphics[width=3in]{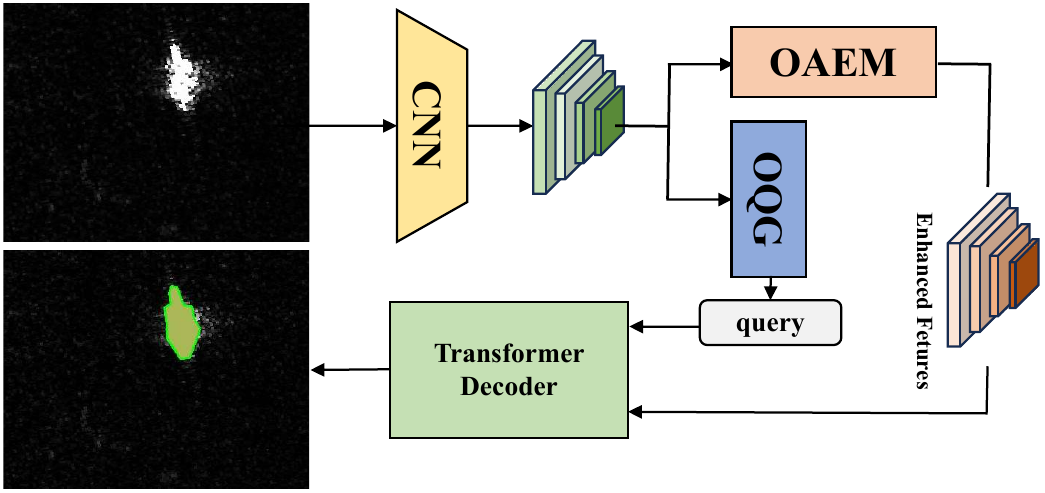}%
\end{wrapfigure}%
\begin{abstract}
Instance segmentation of ships in synthetic aperture radar (SAR) imagery is critical for applications such as maritime monitoring, environmental analysis, and national security. SAR ship images present challenges including scale variation, object density, and fuzzy target boundary, which are often overlooked in existing methods, leading to suboptimal performance. In this work, we propose O\textsuperscript{2}Former, a tailored instance segmentation framework that extends Mask2Former by fully leveraging the structural characteristics of SAR imagery. We introduce two key components. The first is the Optimized Query Generator (OQG). It enables multi-scale feature interaction by jointly encoding shallow positional cues and high-level semantic information. This improves query quality and convergence efficiency. The second component is the Orientation-Aware Embedding Module (OAEM). It enhances directional sensitivity through direction-aware convolution and polar-coordinate encoding. This effectively addresses the challenge of uneven target orientations in SAR scenes.
Together, these modules facilitate precise feature alignment from backbone to decoder and strengthen the model’s capacity to capture fine-grained structural details. Extensive experiments demonstrate that O\textsuperscript{2}Former outperforms state-of-the-art instance segmentation baselines, validating its effectiveness and generalization on SAR ship datasets.
\end{abstract}

\end{minipage}}}

\maketitle

\section{Introduction}
\label{sec:introduction}
\begin{figure}
    \centering
    \includegraphics[width=\linewidth]{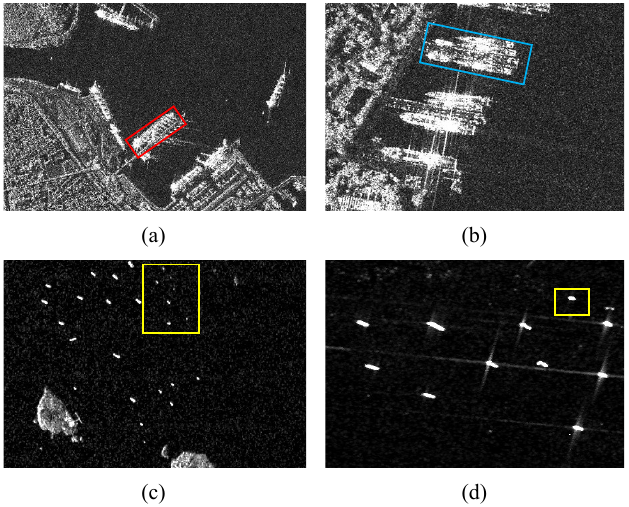}
    \caption{The challenge of ship segmentation in SAR images. (a) shows a scene of direction change. (b) shows a scene of dense distribution. (c) and (d) shows the offshore multi-scale object scene. The ship within the red rectangular box is affected by scattering, resulting in blurred boundaries. The ships in the blue box are too close, complicating boundary segmentation. The small ships in the yellow box are prone to missed detections.}
    \label{fig:ssddexample}
\end{figure}

\IEEEPARstart{S}{ynthetic} Aperture Radar (SAR) plays a crucial role in fishery management, maritime surveillance, and national defense  \cite{mafei}, \cite{langrongling}, \cite{huangheqing}, \cite{10636068} due to its exceptional capability to acquire high-quality imagery under extreme weather conditions, operating effectively both day and night \cite{ma2021fast}, \cite{shan2023sar}. As one of the fundamental applications of SAR imagery, instance segmentation of ships in SAR images has garnered significant attention in recent years \cite{zhongfengjun}, \cite{xu2017gland}, \cite{zheng2021enhancing}. With the advancement of deep learning \cite{zhaowei}, numerous deep neural network frameworks have been widely applied in the field of image segmentation. However, in the context of maritime vessel detection using SAR imagery, several challenging factors complicate accurate segmentation. Fig. \ref{fig:ssddexample} shows the segmentation difficulties in SAR images. These challenges include the diverse scale variations and orientation diversity of target vessels. Furthermore, the imaging of inshore ships is often affected by coastal metallic equipment, making precise segmentation particularly challenging \cite{xu2023edge}, \cite{li2024multiscale}, \cite{li2017improved}. Additionally, vessels in port areas are frequently arranged in compact formations, leading to potential issues such as missed detections and false alarms in these dense scenarios.

Mask2Former \cite{Mask2Former}, developed by Facebook, represents a universal instance segmentation model that employs a set of learnable queries to interact with encoded image features, where each query directly generates an instance. This mask generation approach has demonstrated excellent performance in multi-object scenarios. However, when applied to SAR ship imagery, the model exhibits limitations in detecting small targets and achieving clear boundary segmentation. To address these limitations, this paper aims to design an improved model that better captures and utilizes the unique characteristics of SAR imagery to achieve superior segmentation performance. Our research focuses on enhancing the model's capability to handle the specific challenges posed by SAR ship detection, including scale variations, dense object scenarios, and boundary clarity.

To address these challenges, we propose O\textsuperscript{2}Former, a novel high-resolution SAR image instance segmentation network that integrates an orientation aware embedding module (OAEM) and an optimized query generator (OQG). Our approach significantly improves boundary segmentation accuracy while maintaining robust performance on small targets. Existing query-based instance segmentation methods, such as Mask2Former, rely on zero-initialized queries, which may lead to suboptimal feature alignment between the convolutional backbone and the Transformer decoder. This misalignment can result in inefficient semantic information propagation and slower model convergence. To overcome this limitation, we introduce OQG, which generates query embeddings with rich semantic information by leveraging multi-scale features from the backbone network. This optimization ensures better feature alignment and enhances the model’s ability to capture discriminative ship representations, leading to faster convergence and improved segmentation accuracy.
In addition, the imaging mechanism of SAR images results in a strong correlation between the backscattering characteristics and direction of ship targets. Targets in different directions may exhibit different intensity distributions and shape characteristics. If the network cannot capture changes in directional features, it will be difficult to accurately extract the boundaries and texture details of ships. Ships may be distributed at any angle in the ocean, and SAR images can also introduce angle differences in the imaging direction of ships. The diversity in this direction makes it difficult for traditional method \cite{threshold1}, \cite{edge1}, \cite{region1}, \cite{region2} to capture the global features of the target, which may lead to inconsistent segmentation results. This makes precise instance segmentation particularly challenging. To address these issues, we propose the OAEM. This model can explicitly model the directional information of ship targets. It combines this information with segmentation networks. This approach solves the problem of SAR ship instance segmentation from multiple levels. It enhances the ability to capture target boundaries and directional features. The direction-aware module can extract features of the target in different directions. It dynamically captures important information in the direction by constructing direction-aware convolutions and polar coordinate encoding. 

The main contributions of our research are summarized as follows: 

\begin{itemize}
    
    \item[\textcolor{black}{$\bullet$}] A SAR ship instance segmentation framework has been designed that combines the advantages of orientation information enhancement methods and query based methods. This framework processes multi-scale features through two components, enabling queries to learn instance information at multiple scales which accelerate convergence speed. And the directional information from multiple angles is polarized and encoded, making the model more robust to the directional changes of SAR ship targets.
    \item[\textcolor{black}{$\bullet$}] We have designed an optimized query generator (OQG) that learns efficient semantic information from feature maps by learning a fixed number of query embeddings. This architecture helps queries learn the most informative visual features and generate more accurate masks.
    \item[\textcolor{black}{$\bullet$}] In order to enhance the directional feature extraction capability of the model, we have designed a orientation aware embedding module (OAEM), which can explicitly model the directional features of ships. By using direction aware convolution or direction sensitive attention mechanisms, the network can capture features in different directions, enhance its adaptability to changes in back scattering, and thus improve its robustness to changes in ship direction.
\end{itemize}

The rest of this article is divided into four parts. Section \ref{sec:rw} describes the related work. Section \ref{sec:method} details the methodology of this article. Section \ref{sec:exp} describes the experiments conducted on the polygon segmentation SAR ship detection dataset (SSDD) \cite{SSDD} and high-resolution SAR images dataset (HRSID) \cite{HRSID}. Section \ref{sec:con} concludes this article.

\section{Related Works}
\label{sec:rw}

\subsection{Instance Segmentation}

Convolutional neural network (CNN) based approaches leverage the hierarchical feature extraction capabilities of convolutional networks. These methods \cite{HTC}, \cite{PANet} make improvements based on Mask R-CNN \cite{he2017mask} and feature pyramid network (FPN) \cite{lin2017feature} to make network focus more on small objects and object variations. Mask R-CNN pioneered the integration of region proposal networks with mask prediction, becoming a baseline for SAR ship segmentation. Cascade Mask R-CNN \cite{CascadeRCNN} is an enhanced version of Mask R-CNN that employs a multi-stage approach for instance segmentation. It uses several cascaded detectors to progressively refine predictions, improving accuracy and handling of complex objects. This results in better segmentation quality and robustness in difficult scenarios. MS R-CNN~\cite{MS-RCNN} integrate multi-stage refinement and mask quality learning. Subsequent works like SOLO \cite{Solo} introduced grid-based segmentation to avoid region proposals, while Yoalct \cite{Yolact} combined prototype masks with instance-aware coefficients. For SAR image segmentation, HQ-ISNet \cite{HQISNET} was proposed for remote-sensing image instance segmentation. They have evaluated HQ-ISNet using optical and SAR images but have not considered the characteristic of the targeted SAR ship task, e.g., ship aspect ratio, cross sidelobe, speckle noise, etc. Only generic tricks were offered, still with great obstacles, to further improve accuracy. SISNet \cite{SISNet} focuses on solving scale change problems in remote sensing semantic segmentation tasks. However, due to its insufficient robustness in complex backgrounds, it may misjudge similar texture regions. CATNet~\cite{CATNet} introduces hierarchical context aggregation for remote sensing, while generalist models like SAM~\cite{kirillov2023segment} demonstrate strong transferability via large-scale pretraining.

Transformer-based approaches have emerged as a powerful solution to these challenges by leveraging self-attention mechanisms. The pioneering vision transformer (ViT) \cite{dosovitskiy2020image} laid the foundation for subsequent developments, including the detection transformer (DETR) \cite{DETR}, which opened the era of query based object detection. Building upon this framework, MaskFormer \cite{maskformer} revolutionized segmentation by reformulating per-pixel classification as mask classification. This advancement was further enhanced by Mask2Former \cite{Mask2Former},  which integrated masked attention mechanisms with multi-scale feature representation, achieving remarkable performance in visual recognition tasks. Inspired by these breakthroughs, our study explores the potential of an enhanced query based model for segmenting complex SAR imagery. Segmenter~\cite{Segmenter} and SegFormer~\cite{Segformer}, further boost segmentation performance by leveraging global attention. Recently, remote-sensing-specific methods like RSPrompter~\cite{Rsprompter} and EISP~\cite{EISP} incorporate prompt tuning or hierarchical fusion to address densely distributed and small object segmentation. However, most segmentation models overlook the directional priors inherent in SAR images. Moreover, vision-language pretrained models, such as IMAGGarment-1~\cite{shen2025imaggarment}, have showcased how structured conditioning and layout alignment can boost fine-grained localization and serve as transferable priors for design-aware segmentation tasks.

\subsection{Query-Based Learning}
Transformer queries have redefined detection and segmentation. This type of method introduces a set of queries, which are learnable vectors or representations that can extract information from input data. Each query generates an instance mask, and the model determines the instance corresponding to each query by calculating the similarity between the query and image features.
DETR~\cite{DETR} eliminated anchor design via end-to-end set prediction, followed by variants like DAB-DETR~\cite{dabdetr} and DINO~\cite{dino}, which refine query initialization and convergence. Mask2Former~\cite{Mask2Former} unifies semantic, instance, and panoptic segmentation under a query-driven mask classification framework. Works such as FastInst~\cite{fastinst} show that lightweight decoders and guided queries can balance accuracy and efficiency.
In SAR segmentation, query-based methods remain underexplored. However, query-guided generation has shown promise in related domains. For instance, IMAGDressing-v1~\cite{shen2025imagdressing} employs condition-aware generation through learned queries across vision and text, while IMAGPose~\cite{shen2024imagpose} formulates generation as query-based structure alignment, offering insights into how pretrained queries can drive both segmentation and synthesis tasks. Additionally, consistent query behavior in long-term video generation has been successfully applied in~\cite{shen2025long}, suggesting transferability to temporal segmentation scenarios.

\subsection{Orientation-Aware Modeling}
Orientation-aware modeling is a method used in instance segmentation tasks to enhance the model's understanding of target shape and direction. Explicitly or implicitly modeling the directional characteristics of the target can enable the model to more accurately capture the boundaries and shapes of the target when segmenting directional targets.
Orientation-aware mechanisms have shown notable success in optical tasks like rain removal~\cite{jiang2023dawn} and crowd counting~\cite{kong2023direction}, enhancing robustness to spatial variations. In remote sensing, DiResNet~\cite{chen2021building} and direction-aware attention modules~\cite{cheng2024direction} integrate geometric priors into segmentation. For SAR-specific applications, KL-divergence-based ship detection~\cite{chen2023arbitrary} and directionally guided networks capture ship orientations more effectively.
While SAR imaging is inherently directional due to its side-looking geometry, orientation priors are often ignored in segmentation networks. Inspired by progress in direction-aware vision modeling and structured cross-modal generation~\cite{shen2023advancing}, we aim to embed orientation-awareness directly into feature extraction. Furthermore, leveraging rich contextual priors~\cite{shen2025boosting}, our design bridges visual semantics and directional consistency for complex SAR ship segmentation tasks.

\section{METHODOLOGY}\label{sec:method}

In this section, we will introduce our proposed O\textsuperscript{2}Former, which is a learning approach based on the Mask2Former framework specifically designed for segmentation in SAR images. This section will cover the following aspects: optimizing query generation and orientation aware embedding module.
\begin{figure*}[t]
    \centering
    \includegraphics[width=\textwidth]{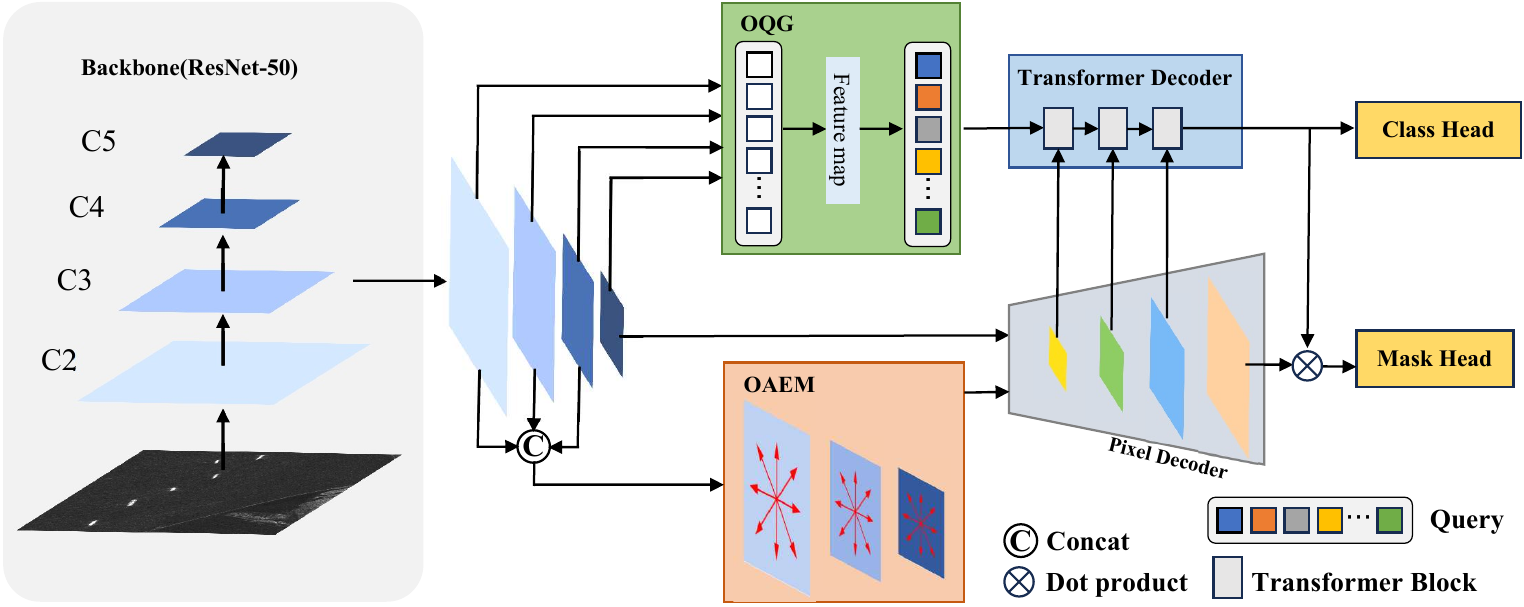}
    \caption{The overall framework of O\textsuperscript{2}Former. The multi-scale features C2, C3, C4 and C5. C2-C5 extracted from SAR images by the CNN backbone network will enter OQG to generate queries. The feature enhancement is performed through OAEM in C2-C4, and the pixel decoder receives the enhanced features and interacts with queries containing multi-scale ship information in the Transformer decoder to generate the final prediction mask.}
    \label{fig:overall architecture}
\end{figure*}

\subsection{Overview}

As illustrated in Fig. \ref{fig:overall architecture}, O\textsuperscript{2}Former consists of five modules: backbone, pixel decoder, optimization query generator, orientation aware module and Transformer decoder. 

Our model feeds an input image \(I \in \mathbb{R}^{H \times W \times 3}\) to the backbone and obtains four feature maps C2, C3, C4, and C5, of which the resolutions are 1/4, 1/8, 1/16, and 1/32 of the input image, respectively. We input four feature maps into the OQG and OAEM, respectively. The pixel decoder aggregates contextual information and outputs enhanced multi-scale feature maps calculated by sparse attention mechanism. It should be noted that A2 output from the pixel decoder does not calculate sparse attention mechanism. It will be added to the upsampled enhanced 1/8 feature map to participate in the calculation of foreground probability.
At the same time, the features output by the backbone network will be mapped to OQG to obtain  \(Q \in \mathbb{R}^{ N \times 256}\) . The transformer decoder takes the total query Q and flattened high-resolution pixel features as inputs, represented as  \(X \in \mathbb{R}^{ L_i  \times 256}\)\((i=1,2,3)\), where \(L_i=H/2^i\times W/2^i \). Then predict the category and segmentation mask at each transformer decoder layer.

\subsection{Optimized Query Generator}
\begin{figure}
    \centering
    \includegraphics[width=1\linewidth]{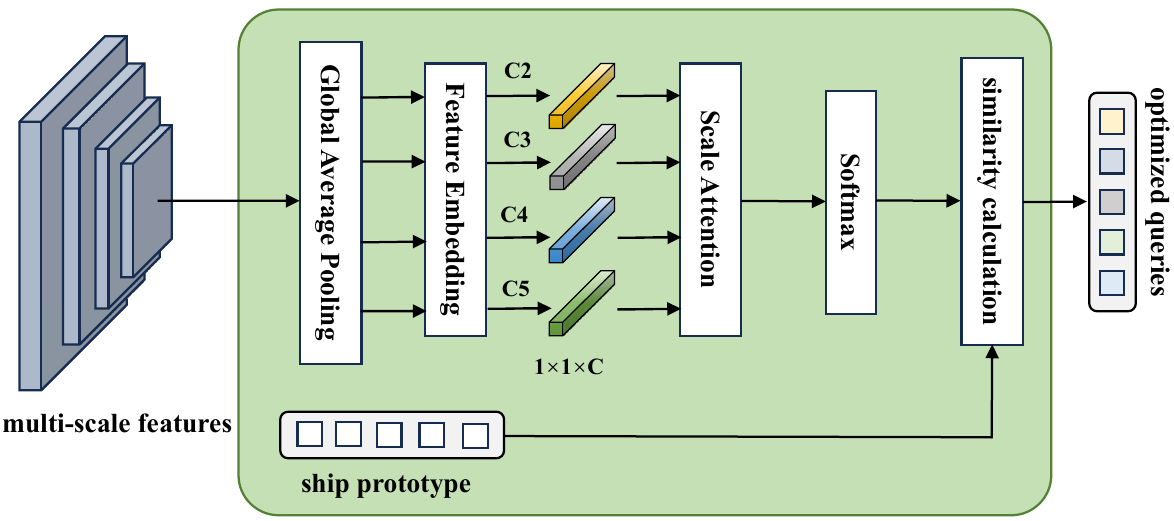}    \caption{The details of the OQG. The multi-scale features are C2, C3, C4 and C5 from the
backbone. Ship prototype is randomly initialized. The optimized queries will generate a mask through encoding feature calculation in Transformer Decoder }
    \label{fig:OQG}
\end{figure}
In instance segmentation tasks, the initialization of queries and feature interaction are key factors that affect model performance. Traditional query initialization methods often lack effective modeling of multi-scale information, resulting in limitations in capturing target features, especially when dealing with scenarios with significant scale changes and complex backgrounds. This deficiency may lead to missed targets and slow model convergence speed. To compensate for the shortcomings of query generation methods, we designed OQG, whose core idea is to integrate multi-scale information into the query initialization process through context aware initialization mechanism, thereby generating query representations containing multi-level feature information.
The detailed structure of OQG is shown in the Fig. \ref{fig:OQG}. A set of randomly initialized ship prototypes and multi-scale features \(X \in \mathbb{R}^{ L_i  \times 256}\)\((i=1,2,3)\), where \(L_i=H/2^i \times  W/2^i \) will be used as inputs for OQG. The channel numbers of the four scale feature maps are 256, 512, 1024, and 2048, respectively. Before inputting into OQG, all channel numbers will be mapped to 256 (not shown in the figure) for subsequent processing. Then, the four feature maps will be globally average pooled to the same size, and be flattened for adding scale embedding in the future. The operation can be formatted as follows: 
\begin{equation}\widetilde{F}_i =Flatten(GlobalAvgPooling(Feature_i))+E_i).\label{eq1}\end{equation}
\begin{equation}\mathcal{F}=Stack([\widetilde{F}_1,\widetilde{F}_2,\widetilde{F}_3,\widetilde{F}_4],dim=1).\label{eq2}\end{equation}

Global average pooling unifies different feature maps to the same size, preserves global information at each scale, and adds scale embedding \(E_i\) to enable the model to distinguish and utilize information at different scales, enhancing feature expression ability.
\(E_i\) is learnable embeddings for each scale and \(Flatten\) is responsible for transforming \(Feature_i\) from a spatial tensor to a vector with a length equal into the number of channels \(C=256\). Afterwards, stack the features to ensure their hierarchical and complete nature. \(\mathcal{F}\in \mathbb{R}^{ B  \times N \times C}\) is the feature after stacking. \(B\) is batch size and \(N=H*W\). Attention weights can ultimately be expressed as follows: 
\begin{equation}
    w_i=softmax(Linear(\mathcal{F})).\label{eq3}
\end{equation}
\begin{equation}
    Feature^*=\sum_{i=0}^{4} w_i\cdot \widetilde{F}_i.\label{eq4}
\end{equation}
Then we use the learnable ship prototype and the encoded multi-scale features to calculate similarity scores, and update the ship prototype with a weight of 0.1 multiplied by this score to obtain an adaptive query. Cosine similarity can be formatted as follows:
\begin{equation}
    S=Feature^*\cdot P^T.\label{eq5}
\end{equation}
The calculation of similarity \(S\) can evaluate the degree of matching between the initial prototype and the fused features, thereby guiding the update direction of the query.
\(P^T\) is initial ship prototype and \(Feature^*\) is weighted fusion features.
The \(Q_{final}\) can be expressed as follows: 
\begin{equation}
    Q_{final}=Linear(P+\eta S^T).\label{eq6}
\end{equation}
\(\eta\) is similarity scaling factor. 
The context aware initialization of OQG used for queries will integrate multi-scale information and ultimately initialize a set of multi-level query representations containing multi-scale feature information. This design can enrich the prior information of queries, enabling better interaction between queries and pixel decoder features, and enhancing the ability of queries to capture features. In addition, using query vectors that have received feature information can also improve the convergence speed of the model.

\subsection{Orientation Aware Embedding Module}
\begin{figure}
    \centering
    \includegraphics[width=1\linewidth]{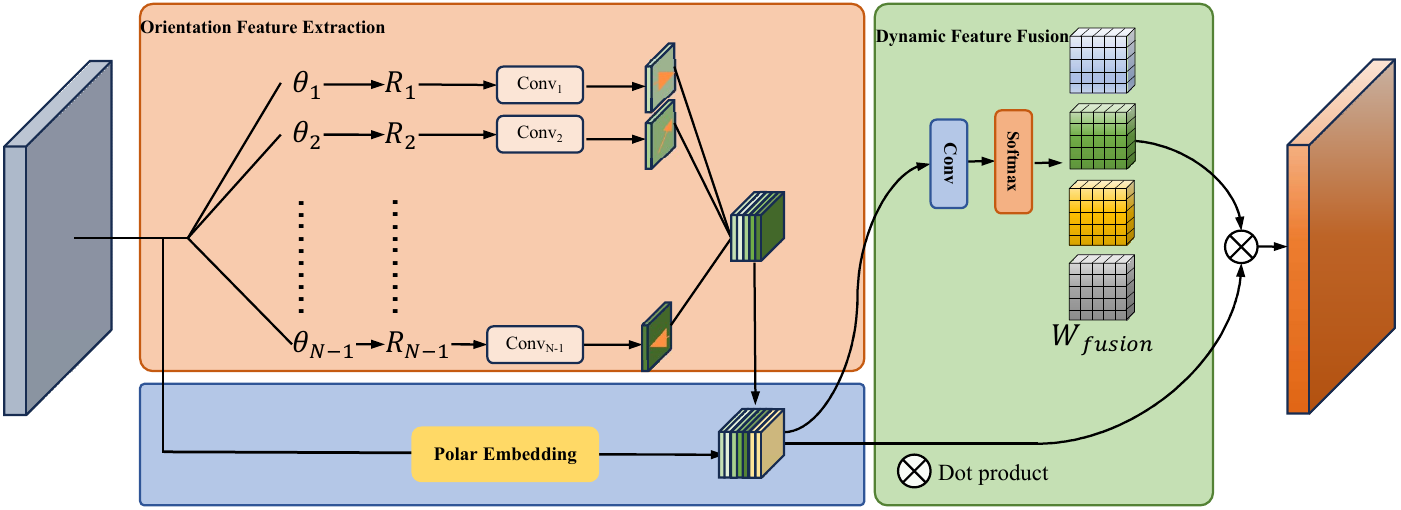}
    \caption{Orientation Aware Embedding Module. $N$ is the preset number of angle convolutions, and the module is divided into three parts: (1) The orientation feature extraction. (2) Polar embedding, (3) Dynamic feature fusion.}
    \label{fig:orientation}
\end{figure}

The diversity of target orientations and the complexity of imaging conditions present significant challenges for achieving accurate instance segmentation in SAR images. This directional diversity complicates the ability of deep learning models to extract stable and robust features. Furthermore, the presence of strong scattering effects and blurred edges exacerbates the difficulty of segmentation tasks, as these phenomena obscure object boundaries and degrade feature clarity.
Existing models, such as Mask2Former, fail to adequately address these challenges when calculating query and input features, particularly in scenarios involving directional diversity. Additionally, prior segmentation approaches have predominantly focused on semantic features during direction-sensitive experiments, often neglecting the critical role of geometric features in representing spatial relationships. This oversight limits the models' ability to fully capture the structural and spatial characteristics necessary for precise segmentation in SAR imagery. We have designed OAEM to address the aforementioned difficulties. 

OAEM mainly consists of three parts: 
1) Orientation feature extraction.
Perform multi angle rotation sampling on input features and extract direction sensitive features through independent angle branch convolution. Each angle corresponds to a dedicated set of convolution kernels that explicitly capture different directional patterns such as edges and textures.
2) Polar embedding. Convert Cartesian coordinates to polar coordinates, encoding the direction and distance information of spatial position. Encode geometric priors of polar coordinates through lightweight convolution.
3) Dynamic feature fusion. Combine direction sensitive features with polar coordinate features and generate the final direction aware embedding through a fusion layer.
The method maintains standard CNN compatibility while enhancing rotation equivariance, as shown in Fig. \ref{fig:orientation}.

Given input feature map   \(X \in \mathbb{R}^{ C  \times H \times W}\). For \(N\) predefined angles \(\theta_i = \frac{i\pi}{N}, i=0,1,..., N-1\) :
\begin{equation}
    {X}_{i}^{{rot}} = {GridSample}({X}, {AffineGrid}({R}_{i}))\in \mathbb{R}^{(C/N)\times H \times W},\label{eq7}
\end{equation}
\(R_i\) is the rotation matrix for \(\theta_i\). AffineGrid is used to generate a rotated sampling grid. \(X_i^{rot}\) explicitly encodes the local structure of input features in terms of angles. Next, we use parallel convolution branches to process the rotation feature \(X_i^{rot}\). 
\begin{equation}
    F_i=Conv_i(X_i^{rot}), i=0,...,N-1.\label{eq8}
\end{equation}
\(Conv_i\) is the convolution operation corresponding to the angle and the dimension remains unchanged. The final feature with angle information will be represented as 
\begin{equation}
    F_{orient}=Concat(F_0,F_1,...,F_{N-1}))\in \mathbb{R^{C \times H \times W}}.\label{eqx}
\end{equation}

In order to encode directional features in polar coordinates, it is necessary to convert the Cartesian coordinate system into a polar coordinate representation: 
\begin{equation}
    \begin{cases}
r = \sqrt{x^2 + y^2} \\
\theta = \operatorname{atan2}(y, x)
\end{cases}.\label{eq9}
\end{equation}

For numerical stability, we normalize the polar coordinates

\begin{equation}
    \begin{cases}
r_{norm} = r/\sqrt{2} \in[0,1]\\
\theta_{norm} = \frac{\theta+\pi}{2\pi}\in[0,1]
\end{cases},\label{eq10}
\end{equation}
and stack the normalized \((r,\theta)\) to form a new tensor \(F_{polar}\), which simply encodes geometric properties.
\begin{equation}
    F_{polar}=Stack(r_{norm}, \theta_{norm})\in \mathbb{R}^{2 \times H \times W}.\label{eq11}
\end{equation}
The \(\theta\) of polar coordinates directly encodes directional information, allowing the convolution kernel to explicitly learn rotation patterns.
When an object rotates, its polar coordinates only change by \(\theta\), while r remains unchanged. The radius r distinguishes the central and edge regions of the feature map, making it suitable for detecting targets of different scales. 
Finally, dynamic feature fusion is performed, which weights direction sensitive features and polar coordinate features by generating weights \(W_{fusion}\).Firstly, concatenate the direction sensitive features and polar coordinate features.
\begin{equation}
    F_{concat}=Concat(F_{orient},F_{polar}).
\end{equation}
Generate fusion weights through dynamic fusion module:
\begin{equation}
    W_{fusion}=Softmax(Conv(F_{concat})).
\end{equation}
Finally, use weights to weight and fuse the two features.
\begin{equation}
    F_{fusion}=F_{orient} \odot W_{fusion}+F_{poalr} \odot (1-W_{fusion}).
\end{equation}
$\odot$ is element wise multiplication.
The integration of directional sensitive features with polar coordinates offers a powerful approach to enhancing the robustness of ship instance segmentation in SAR images. Directional sensitive convolutions capture semantic features that are crucial for understanding the shapes and patterns of ships, while polar coordinates provide essential geometric priors that help in accurately representing spatial relationships. Dynamic feature fusion enables the model to adaptively select the contributions of direction sensitive features and polar coordinate features during feature fusion, thereby improving the performance of the model in handling different targets and scenes. This dynamic weighting mechanism is the key to achieving efficient feature fusion.By combining these methodologies, the module effectively addresses the challenges posed by varying orientations in SAR imagery. The orientation-sensitive processing allows the model to adapt to different ship headings, ensuring that it can recognize and segment ships regardless of their position or angle in the image. Meanwhile, the polar coordinate transformation enriches the feature representation by encoding direction and distance information, which is vital for distinguishing between closely situated objects. The dynamic feature fusion process further enhances the model's performance by integrating the semantic and geometric features, resulting in a more comprehensive understanding of the scene. This synergy not only improves the accuracy of ship instance segmentation but also increases the model's generalizability across diverse scenarios. In summary, the combination of orientation-sensitive processing, polar coordinate transformation, and dynamic feature fusion creates a robust framework for ship instance segmentation in SAR images, effectively overcoming the limitations associated with directional diversity and enhancing overall model performance.  

\section{EXPERIMENTS}\label{sec:exp}

\subsection{Datasets}
To validate the proposed O\textsuperscript{2}Former method's superiority, it is compared with multiple state-of-the-art instance segmentation approaches on two SAR ship datasets, namely SSDD and HRSID.
The SSDD \cite{SSDD} dataset is a publicly available SAR ship image dataset designed. Due to subsequent extensions by researchers, it is also referred to as PSeg-SSDD or SSDD++. The dataset consists of 1,160 ocean SAR images with resolutions ranging from 1 to 15 meters. These images were captured by different sensor models and polarizations, containing a total of 2,587 ships. In our experiments, we followed the official release standards of PSeg-SSDD, dividing the dataset into 928 training images and 232 test images. The test set includes 46 images from offshore scenes and 186 images from nearshore scenes.

The HRSID \cite{HRSID} dataset consists of 138 panoramic SAR images with resolutions ranging from 1 to 5 meters. Additionally, it includes 5,605 image slices, each with a resolution of 800×800 pixels, representing diverse imaging modes, polarization techniques, and resolutions. The dataset contains annotations for 16,951 ships, with a balanced distribution of object sizes: 9,242 small targets, 7,388 medium-sized targets, and 321 large targets. Following the official release standards of HRSID, we divided the dataset into 3,642 training images and 1,962 test images to evaluate the performance of each model across various scenarios. 
\subsection{Evaluation Metric}

Microsoft Common Objects in Context (MS COCO) evaluation metric \cite{coco} is the most common performance evaluation metric in the field of instance segmentation. The core of  MS COCO is the intersection ratio of the mask prediction to the mask ground truth. Based on the IoU threshold, the precision and recall of the target can be calculated

\begin{equation}
    presion = \frac{TP}{TP + FP},\label{eq12}
\end{equation}
and
\begin{equation}
    recall = \frac{TP}{TP + FN},\label{eq13}
\end{equation}
where \(TP\) is the number of true positives, \(FP\) is the number of false positives, and \(FN\) is the number of false negatives. According to precision and recall, the average precision (\(AP\)) can be calculated as follows:
\begin{equation}
    AP = \int_{0}^{1} p(r)dr.\label{eq14}
\end{equation}

Here \(r\) represents the recall rate and \(p(r)\) represents the precision when recall rate is \(r\).  The mean average precision (\(mAP\)) represents the \(AP\) of multiple categories. In the field of SAR ship detection, there is only one category, which is the ship, so \(AP\) and \(mAP\) are equal.

\begin{table}[t]
\centering
\caption{MS COCO METRICS}
\label{tab:coco_metrics}
\renewcommand{\arraystretch}{1.5}
\begin{tabular}{lp{6cm}} 
\toprule
\textbf{Metrics} & \textbf{Meaning} \\
\midrule
$AP_m$ & IoU=0.50:0.05:0.95 \\
$AP_m^{50}$  & IoU=0.50 \\
$AP_m^{75}$  & IoU=0.75 \\
$AP_m^{S}$   & AP of small objects: area \(< 32^2\) \\
$AP_m^{M}$   & AP of medium objects: \(32^2 < \text{area} < 64^2\) \\
$AP_m^{L}$   & AP of large objects: area \(> 64^2\) \\
\bottomrule
\end{tabular}
\end{table}

According to different IoU thresholds and object pixel areas, MS COCO has various evaluation metrics to comprehensively evaluate the accuracy of object detection. Table \ref{tab:coco_metrics} lists these evaluation metric.

\begin{table*}[t]
    \centering
    \renewcommand{\arraystretch}{1.5}
    \caption{INSTANCE SEGMENTATION PERFORMANCE OF DIFFERENT METHODS ON THE PSeg-SSDD.}
    \label{tab:performance1}
    \begin{tabularx}{\textwidth}{lXXXXXXXXXXXX}
        \hline
        \multirow{2}{*}{\textbf{Method}} & \multicolumn{6}{c}{\textbf{Offshore scene (\%)}} & \multicolumn{6}{c}{\textbf{Inshore scene (\%)}} \\ \cline{2-7} \cline{8-13}
        & \textbf{AP$_m$} & \textbf{AP$_m^{50}$} & \textbf{AP$_m^{75}$} & \textbf{AP$_m^{S}$} & \textbf{AP$_m^{M}$} & \textbf{AP$_m^{L}$} & \textbf{AP$_m$} & \textbf{AP$_m^{50}$} & \textbf{AP$_m^{75}$} & \textbf{AP$_m^{S}$} & \textbf{AP$_m^{M}$} & \textbf{AP$_m^{L}$}\\ \hline
        Mask R-CNN \cite{he2017mask} & 64.4 & 97.9 & 78.6 & 64.7 & 64.5 & 70.0 & 41.9 & 68.8 & 47.4 & 47.4 & 29.8 & 14.2 \\ 
        Cascade Mask R-CNN \cite{CascadeRCNN} & 64.6 & 98.5 & 79.1 & 64.3 & 66.8 & 70.0 & 45.1 & 73.9 & 52.2 & 49.5 & 35.3 & 50.0 \\ 
        Yolcat \cite{Yolact} & 61.4 & 95.8 & 75.2 & 61.6 & 62.4 & 70.0 & 35.3 & 61.0 & 40.4 & 39.9 & 25.8 & 4.1 \\ 
        CondInst \cite{tian2020condistl} & 58.0 & 97.3 & 68.2 & 54.2 & 72.0 & 60.0 & 42.1 & 83.0 & 36.5 & 39.2 & 49.0 & 90.0 \\ 
        RTMDet \cite{rtmdet} & 60.3 & 98.1 & 73.8 & 57.1 & 71.7 & 70.0 & 41.5 & 83.0 & 36.0 & 34.2 & 60.3 & 40.0 \\
        HQ-ISNet \cite{HQISNET} & 66.6 & 98.6 & 82.0 & 65.5 & 70.5 & 28.3 & 45.5 & 75.9 & 52.9 & 45.3 & 47.4 & 40.0 \\
        AFSS-Inst \cite{AFSS} & 65.9 & 98.6 & 82.7 & 63.4 & 75.1 & 60.0 & 45.5 & 81.5 & 47.7 & 45.2 & 47.7 & 36.7 \\
        SRNet \cite{srnet} & 66.1 & 97.9 & 82.8 & 64.0 & 73.7 & 70.0 & 44.2 & 76.6 & 47.7 & 44.0 & 45.3 & 60.0 \\
        Mask2Former \cite{Mask2Former} & 67.2 & 94.0 & 83.2 & 65.3 & 77.8 & 60.0 & 52.5 & 83.0 & 59.9 & 51.4 & 56.7 & 90.0 \\
        DiffSARShipInst \cite{diffsarshipinst} & 70.6 & \textbf{98.8} & 90.8 & 69.1 & 78.1 & 60.0 & 56.2 & 89.7 & 67.1 & 53.4 & 64.4 & 90.0 \\
        \textbf{O\textsuperscript{2}Former (Ours)} & \textbf{71.9} & 98.0 & \textbf{92.9} & \textbf{70.8} & \textbf{78.7} & \textbf{90.0} & \textbf{63.2} & \textbf{91.6} & \textbf{81.2} & \textbf{59.5} & \textbf{73.9} & \textbf{90.0} \\ \hline
    \end{tabularx}
\end{table*}

\begin{table*}[t]
    \centering
    \renewcommand{\arraystretch}{1.5}
    \caption{INSTANCE SEGMENTATION PERFORMANCE OF DIFFERENT METHODS ON THE HRSID.}
    \label{tab:performance2}
    \begin{tabularx}{\textwidth}{lXXXXXXXXXXXX}
        \hline
        \multirow{2}{*}{\textbf{Method}} & \multicolumn{6}{c}{\textbf{Offshore scene (\%)}} & \multicolumn{6}{c}{\textbf{Inshore scene (\%)}} \\ \cline{2-7} \cline{8-13}
        & \textbf{AP$_m$} & \textbf{AP$_m^{50}$} & \textbf{AP$_m^{75}$} & \textbf{AP$_m^{S}$} & \textbf{AP$_m^{M}$} & \textbf{AP$_m^{L}$} & \textbf{AP$_m$} & \textbf{AP$_m^{50}$} & \textbf{AP$_m^{75}$} & \textbf{AP$_m^{S}$} & \textbf{AP$_m^{M}$} & \textbf{AP$_m^{L}$}\\ \hline
        Mask R-CNN \cite{he2017mask} & 65.3 & 95.8 & 83.4 & 64.5 & 72.5 & 50.0 & 34.0 & 56.5 & 38.3 & 31.0 & 54.9 & 33.4 \\ 
        Cascade Mask R-CNN \cite{CascadeRCNN} & 67.4 & 96.8 & 85.7 & 66.7 & 74.3 & 41.3 & 35.6 & 59.1 & 40.1 & 32.6 & 55.6 & 33.6 \\ 
        Yolcat \cite{Yolact} & 58.6 & 95.6 & 71.0 & 58.0 & 68.3 & 35.6 & 27.2 & 61.8 & 18.7 & 25.8 & 41.9 & 9.6 \\ 
        CondInst \cite{tian2020condistl} & 60.7 & 97.6 & 74.8 & 59.7 & 70.3 & 61.3 & 32.0 & 71.6 & 24.2 & 30.0 & 50.3 & 13.5 \\ 
        RTMDet \cite{rtmdet} & 67.4 & 97.7 & 83.7 & 66.8 & 74.9 & 33.6  & 40.4 & 75.9 & 38.7 & 39.3 & 53.0 & 10.1 \\
        HQ-ISNet \cite{HQISNET} & 67.9 & 97.5 & 87.1 & 66.5 & 73.8 & 46.8 & 39.6 & 72.5 & 41.0 & 39.0 & 49.7 & 9.5 \\
        AFSS-Inst \cite{AFSS} & 65.2 & 96.4 & 83.4 & 67.3 & 61.1 & 10.3 & 28.2 & 51.8 & 28.0 & 28.0 & 33.5 & 3.1 \\
        SRNet \cite{srnet} & 67.7 & 96.9 & 86.0 & 74.9 & 59.5 & 22.3 & 39.5 & 71.6 & 40.6 & 16.2 & 14.7 & 8.2 \\
        Mask2Former \cite{Mask2Former} & 60.6 & 95.2 & 74.9 & 59.5 & 73.2 & 73.2 & 17.9 & 36.9 & 16.2 & 14.7 & 41.8 & 31.1 \\
        DiffSARShipInst \cite{diffsarshipinst} & 70.9 & \textbf{97.8} & 90.0 & 70.3 & 77.5 & 73.2 & 42.6 & \textbf{76.1} & 44.1 & \textbf{40.3} & 58.7 & 43.6. \\
        \textbf{O\textsuperscript{2}Former (Ours)} & \textbf{71.7} & 97.5 & \textbf{91.0} & \textbf{71.1} & \textbf{79.4} & \textbf{74.9} & \textbf{43.1} & 69.5 & \textbf{45.9} & 39.9 & \textbf{58.8} & \textbf{50.6} \\ \hline
    \end{tabularx}
\end{table*}

\subsection{Implementation Details}

The proposed method was tested on SSDD as well as HRSID, initialized on HRSID using ResNet-50 pretrained weights on the ImageNet dataset and fine-tuned on SSDD using weights on the HRSID. The experiments are implemented on PyTorch 3.8 and CUDA 12.2 with Nvidia Geforce RTX 4090 GPU. The dataset is partitioned using random sampling and the batch size is set to 8. The learning rate decreasing method is used with the multi step function. The optimizer is AdamW. Data enhancement methods such as random flip, random scale, and cropping are used, and the initial learning rate is set to 0.0001. The training epoch is 500 and learning rate reduced by a factor of 10 at epochs 300 and 400.

\subsection{Experimental Results}
To verify the effectiveness and robustness of our method, we implement experiments on two datasets of PSeg-SSDD and HRSID. \(AP_m\), \(AP_m^{50}\), \(AP_m^{75}\), \(AP_m^{S}\), \(AP_m^{M}\), and \(AP_m^{L}\) are used to evaluate the performance of different methods, i.e., Mask R-CNN \cite{he2017mask}, Cascade Mask R-CNN \cite{CascadeRCNN}, Mask Scoring R-CNN \cite{MS-RCNN}, Yolact \cite{Yolact}, CondInst \cite{tian2020condistl}, RTMDet \cite{rtmdet}, HQ-ISNet \cite{HQISNET}, AFSS-Inst \cite{AFSS}, SRNet \cite{srnet}, Mask2Former \cite{Mask2Former} and DiffSARShipInst \cite{diffsarshipinst}. We conduct our experiments according to the division of inshore and offshore scenes. Table \ref{tab:performance1} and Table \ref{tab:performance2} demonstrates the quantitative performance of different methods. 
Our proposed O\textsuperscript{2}Former demonstrates superior performance in ship instance segmentation across both the PSeg-SSDD and HRSID datasets, consistently outperforming existing state-of-the-art methods in most evaluation metrics. An analysis of the experimental results is provided as follows. 

\subsubsection{Comparisons on PSeg-SSDD}   
O\textsuperscript{2}Former has achieved significant improvement in instance segmentation performance on the PSeg SSDD dataset and has significant advantages compared to existing methods. As shown in Table \ref{tab:performance1}, in the offshore scenario, the  \(AP_m\) of O\textsuperscript{2}Former is 71.9\%, which is 7.5\%, 7.3\%, 10.5\%, 13.9\%, 11.6\%, 5.3\%, 6.0\%, 5.8\%, 4.7\%, and 1.3\% higher than the previous methods Mask R-CNN, Cascade R-CNN, Yolcat, CondInst, RTMDet, HQ-ISNet, ASFF-Inst, SRNet, Mask2Former and DiffSARShipInst on the \(AP_m\), respectively. For inshore scenario, this improvement is more pronounced, with O\textsuperscript{2}Former achieving an  \(AP_m\) of 63.2\%, significantly better than all previous methods, and 7.0\% higher than the highest performance DiffSARShipInst (56.2\%). Compared to the baseline Mask2Former, adding OQG and OAEM modules has greatly improved its performance in inshore scenarios. In \(AP_m\), \(AP_m^{50}\) and \(AP_m^{75}\) , there have been improvements of 10.7\%, 8.3\% and 21.3\% respectively. This is because OQG allows queries to learn the features of ships of different sizes before entering the Transformer decoder, enabling better information exchange with pixel features and optimizing segmentation performance. At the same time, OAEM can autonomously perceive different angles of the ship, which can to some extent suppress the impact of scattered noise and better complete the task in the case of uneven distribution of ship angles.
\begin{figure}[t]
    \centering
    \includegraphics[width=\linewidth]{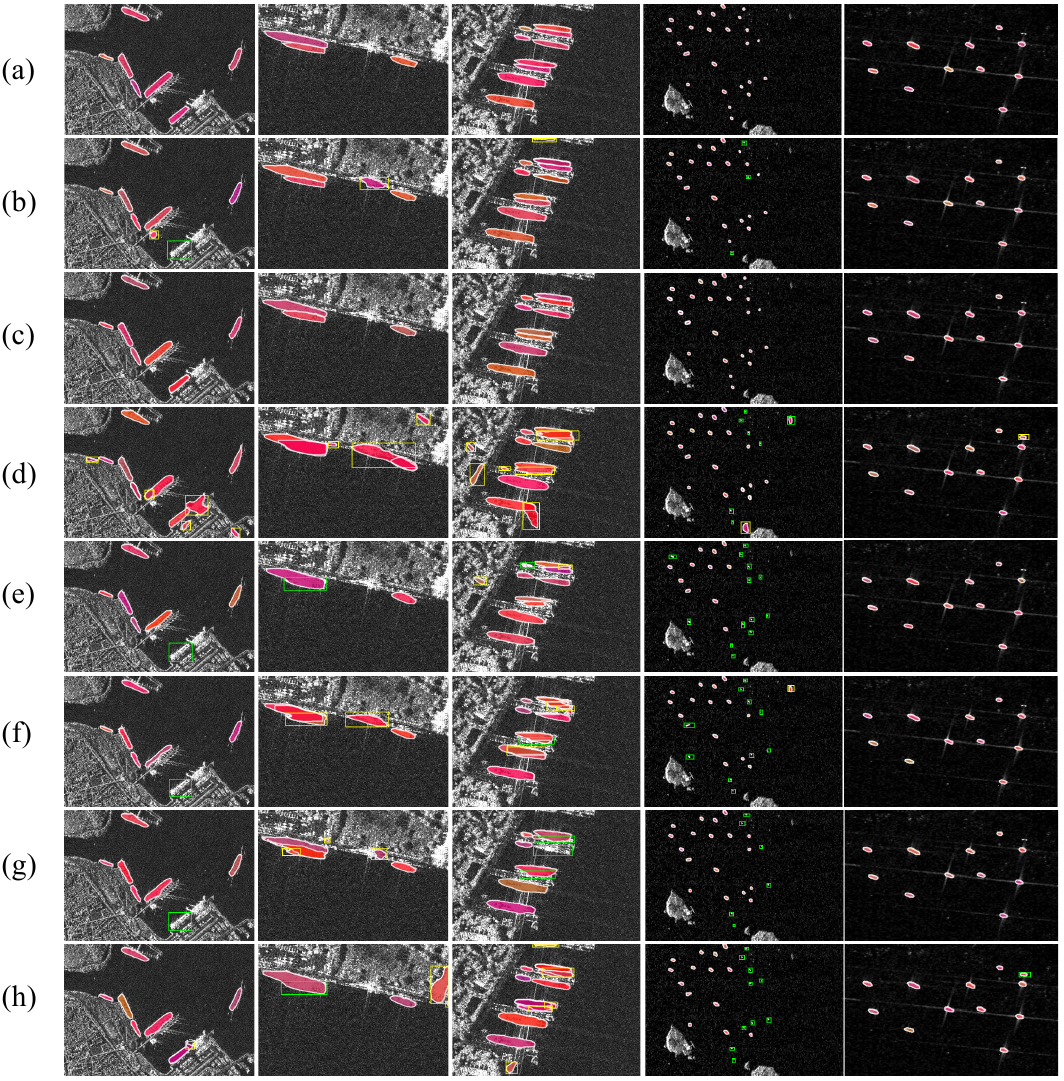}
    \caption{Instance segmentation results of different methods on PSeg-SSDD. (a) Ground truth. (b) Mask2Former. (c) Our method. (d) Mask RCNN. (e) Cascade Mask RCNN. (f) Yolact. (g) CondInst. (h) RTMDet. The yellow rectangular box represents false alarms, and the green rectangular box represents missed detections.}
    \label{fig:visionssdd}
\end{figure}
\subsubsection{Comparisons on HRSID}    
HRSID has a more complex background clutter and a large number of small ships, making it more challenging. But the experiments on the HRSID dataset still validate O\textsuperscript{2}Former's effectiveness. In the offshore scenario, the  \(AP_m\) of O\textsuperscript{2}Former is 71.7\%, which is 6.4\%, 4.3\%, 13.1\%, 11.0\%, 4.3\%, 3.8\%, 6.5\%, 4.0\%, 11.1\%, and 0.8\% higher than the previous methods Mask R-CNN, Cascade R-CNN, Yolcat, CondInst, RTMDet, HQ-ISNet, ASFF-Inst, SRNet, Mask2Former and DiffSARShipInst on the AP, respectively. Except for \(AP_{m}^{50}\) which is slightly inferior to DiffSARShipInst, all others have the highest performance. For inshore environments, our method achieved a performance of  45.9\% under high IoU threshold \(AP_{m}^{75}\), leading all methods in the table. And the segmentation effect for medium and large targets has also reached SOTA. This indicates that O\textsuperscript{2}Former can still generate high-quality prediction masks on challenging datasets such as HRSID. This is attributed to the directional convolution introduced in OAEM, which enables the model to perceive the directional features of boundaries more clearly.
\begin{figure}[t]
    \centering
    \includegraphics[width=0.98\linewidth]{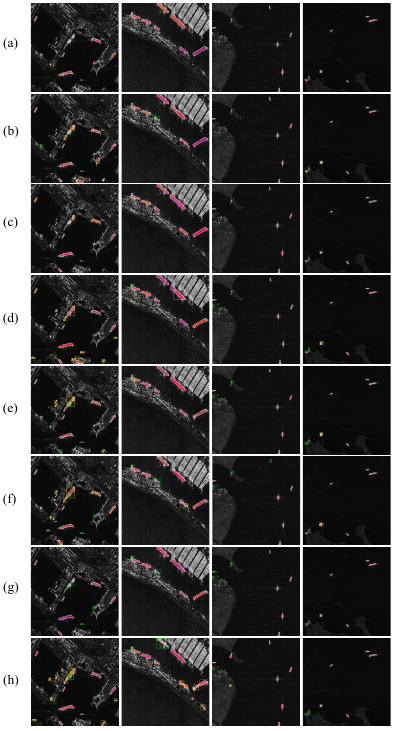}
    \caption{Instance segmentation results of different methods on HRSID. (a) Ground truth. (b) Mask2Former. (c) Our method. (d) Mask RCNN. (e) Cascade Mask RCNN. (f) Yolact. (g) CondInst. (h) RTMDet. The yellow rectangular box represents false alarms, and the green rectangular box represents missed detections.}
    \label{fig:visionhrsid}
\end{figure}

Fig. \ref{fig:visionssdd} presents the visualization results of Mask2Former, O\textsuperscript{2}Former, Mask R-CNN, Cascade Mask R-CNN, Yolact, CondInst, and RTMDet on the PSeg-SSDD dataset. As shown in Fig. \ref{fig:visionssdd}, Cascade Mask R-CNN and Yolact exhibit significant omissions in detecting small offshore targets, while Mask R-CNN, Yolact, and RTMDet are prone to false positives when processing nearshore targets. In contrast, O\textsuperscript{2}Former outperforms other methods in handling small targets, which can be attributed to the multi-scale feature fusion capability of the OQG module. By integrating multi-level global and local information into the query generation process, OQG enriches the feature representation, enabling the model to capture fine-grained details of small offshore targets and significantly reducing missed detections. From the perspective of fine-grained segmentation, Mask R-CNN and CondInst are affected by the scattering characteristics of SAR images, leading to over-segmentation of targets. The OAEM module, through its orientation-sensitive convolution, extracts features containing directional information, allowing the model to better adapt to ships with varying orientations, especially in offshore scenarios with diverse directional distributions. Compared to other methods, O\textsuperscript{2}Former demonstrates higher robustness in handling targets with varying orientations in offshore scenes. Additionally, as shown in the second and third images of Fig. \ref{fig:visionssdd} (c), O\textsuperscript{2}Former excels in segmenting closely located targets. This is primarily due to the following reasons: OQG provides rich multi-scale features and contextual information, which serve as high-quality input for OAEM. OAEM further optimizes the queries generated by OQG through orientation-aware and geometric modeling, enabling the queries to better adapt to directional diversity and complex scenarios. The combination of the two modules achieves dynamic fusion of semantic and geometric features, forming a comprehensive feature representation framework for O\textsuperscript{2}Former.

Fig. \ref{fig:visionhrsid} presents the visualization results of Mask2Former, O\textsuperscript{2}Former, Mask R-CNN, Cascade Mask R-CNN, Yolact, CondInst, and RTMDet on the HRSID dataset. As shown in Figure \ref{fig:visionhrsid}, Mask R-CNN, Cascade Mask R-CNN, and RTMDet are prone to false positives in complex backgrounds, where some background regions are mistakenly segmented as targets. Additionally, Mask2Former and RTMDet exhibit limited capability in representing target boundaries, with some targets being segmented into multiple instances. In the second column of Fig. \ref{fig:visionhrsid}, only O\textsuperscript{2}Former successfully detects the small target at the stern on the far left, while all other methods fail, resulting in missed detections. This success is attributed to the orientation-sensitive features introduced by OAEM, which help the model more accurately capture the boundary information of closely packed targets, thereby reducing target adhesion. The polar coordinate geometric priors further enhance the model's ability to distinguish targets from the background, resulting in more precise boundary representations in complex backgrounds. O\textsuperscript{2}Former significantly reduces missed detections of small targets in offshore scenarios, producing more complete segmentation results. In nearshore scenarios, O\textsuperscript{2}Former effectively reduces false positives, while also addressing issues of target adhesion and over-segmentation. The combination of OQG and OAEM provides complementary semantic and geometric features, enabling the model to achieve stronger robustness and generalization in complex scenarios.

\begin{table}[t]
    \centering
    \renewcommand{\arraystretch}{1.5}
    \caption{Results of Ablation Experiments on the SSDD}
    \begin{tabularx}{0.5\textwidth}{XXXXXXXXX}
        \hline
        OQG & OAEM & Scene & AP & AP$_{50}$ & AP$_{75}$ & AP$_{S}$ & AP$_{M}$ & AP$_{L}$ \\
        \hline
        \multirow{2}{*}{\text{-}} & \multirow{2}{*}{\text{-}} & offshore & 0.672 & 0.940 & 0.832 & 0.653 & 0.778 & 0.600  \\
        & & inshore & 0.525 & 0.830 & 0.599 & 0.514 & 0.567 & 0.900 \\
        \hline
        \multirow{2}{*}{\checkmark} & \multirow{2}{*}{\text{-}} & offshore & 0.695& 0.965 & 0.896 & 0.679 & 0.781 & 0.700 \\
        & & inshore & 0.556 & 0.851 & 0.647 & 0.515 & 0.683 & 0.900 \\
        \hline
        \multirow{2}{*}{\text{-}} & \multirow{2}{*}{\checkmark} & offshore & 0.710 & 0.962 & \textbf{0.930} & 0.698 & 0.783 & 0.600 \\
        & & inshore & 0.552 & 0.851 & 0.631 & 0.540 & 0.606 & 0.900 \\
        \hline
        \multirow{2}{*}{\checkmark} & \multirow{2}{*}{\checkmark} & offshore & \textbf{0.719} & \textbf{0.980} & 0.929 & \textbf{0.708} & \textbf{0.787} & \textbf{0.900} \\
        & & inshore & \textbf{0.632} & \textbf{0.916} & \textbf{0.812} & \textbf{0.595} & \textbf{0.739} & \textbf{0.900} \\
        \hline
    \end{tabularx}
    \label{tab:ssdd_ablation_results}
\end{table}

\subsection{Ablation Experiment}

To demonstrate the effectiveness of each part proposed in this article and quantitatively evaluate its improvement, we conducted ablation experiments on SSDD using OQG and OAEM modules. Tables \ref{tab:ssdd_ablation_results} show the results of the OQG module and OAEM ablation experiments, respectively. By comparing the model maps, we can draw the following conclusions.

The ablation experiment results in Table \ref{tab:ssdd_ablation_results} indicate that the independent application of OQG leads to a 2.3\% and 3.1\% increase in \(AP_m\) for both offshore and nearshore areas, respectively. Meanwhile, the independent application of OAEM will increase \(AP_m\) by 4.8\% and 2.7\% for both offshore and nearshore areas, respectively. It can be seen that the two modules mainly improve the segmentation accuracy of medium and small targets in the PSeg-SSDD dataset. During the process of generating queries, OQG continuously learns multi-scale information from features, enhancing the initial query's perception of targets of different sizes. It has achieved 2.6\%, 0.3\%, and 10.0\% improvements in far-field scenarios for \(AP_m^{S}\), \(AP_m^{M}\), and \(AP_m^{L}\). OQG increased \(AP_m\) by 11.6\% in nearshore scenarios. OAEM focuses on extracting directional information, including ship orientation and target scattering characteristics. There were 4.5\% and 0.5\% improvements in far-field scenarios and 2.6\% and 3.9\% improvements in nearshore scenarios for \(AP_m^{S}\) and \(AP_m^{M}\), respectively.
Due to the extremely small proportion of large targets in the PSeg-SSDD dataset. The number of large targets in nearshore and offshore scenarios is 19 and 52, respectively, accounting for only 3.58\% and 2.54\% of the total number. Therefore, the imbalance of data results in the two modules allocating more attention to small and medium-sized targets, making it extremely difficult to improve the enhancement of large targets due to data distribution issues. But overall, both modules have made significant contributions to improving network accuracy.
By comparing the improvements of the two modules, it can be observed that OQG and OAEM have improved \(AP_m^{75}\) by 6.4\% and 9.8\% respectively under the condition of high IOU threshold on the far shore, indicating that the insertion of OQG enhances the ability to capture global contextual information, while OAEM enhances the ability to capture boundary details. The two modules start from both global and local perspectives, complementing each other to further improve the segmentation performance of the network.
\begin{figure}[t]
    \includegraphics[width=0.5\textwidth]{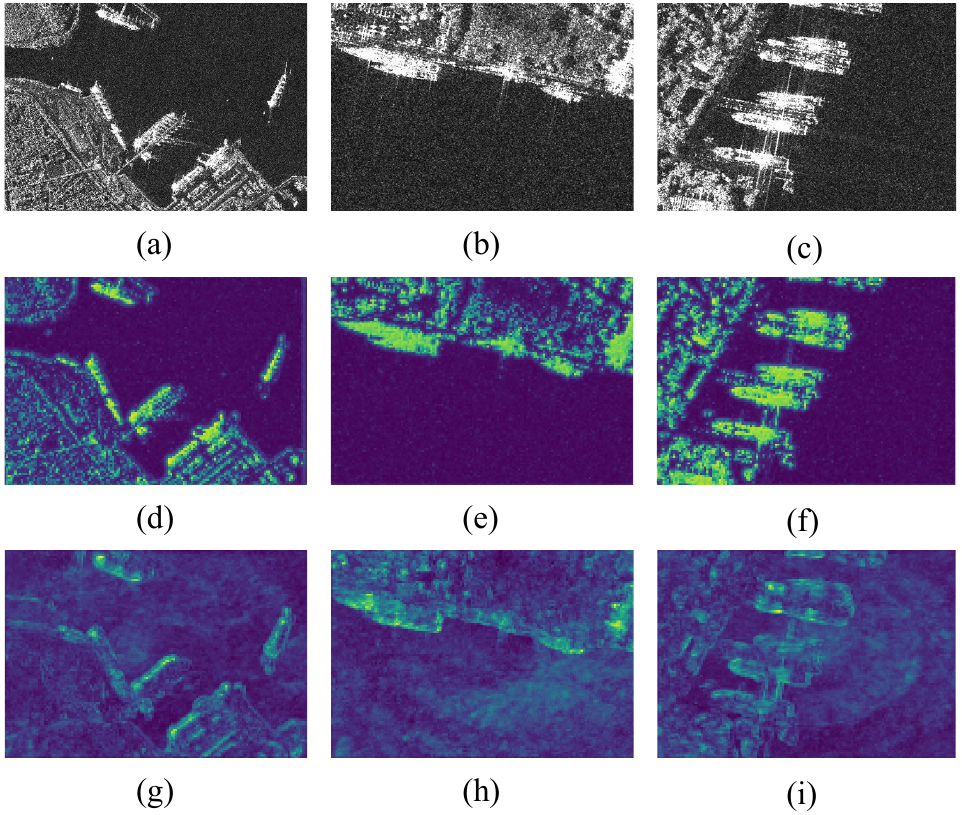}
    \caption{Visualization of directional feature enhancement results. The first line represents the original SAR image, the second line represents the feature map enhanced by OAEM, and the third line represents the feature map without OAEM. (d) Enhancement effects in scenes with the multi orientation distribution of ships. (e) Enhancement effects in scenes with complex environmental backgrounds. (f) Enhancement effects in scenes with the dense distribution of ships. }
    \label{fig:ssddvisfeature}
\end{figure}

In order to verify the effectiveness of OAEM more clearly and intuitively, this paper chose the shallower feature \(C2\) output from the backbone network for visualization. The specific results are shown in Fig. \ref{fig:ssddvisfeature}. It covers three representative scenarios: 

1. Multi directional distribution of ship scenes.
In the feature map without OAEM (Fig. \ref{fig:ssddvisfeature} (g)), the model extracts blurry features for ships in different directions, especially for important directional structures such as bow and stern, which are difficult to express clearly. The boundary information of the ships is also incomplete, and some ships in certain directions have even experienced severe feature loss.
After adding OAEM (Fig. \ref{fig:ssddvisfeature} (d)), the ship boundaries in the feature map became clearer, especially for ships in different directions, and their structural features were significantly enhanced. Direction sensitive convolution effectively captures the directional features of ships, enabling the model to more accurately identify and segment targets with different orientations.

2. Complex coastal environment background scene.
In the feature map without OAEM (Fig. \ref{fig:ssddvisfeature} (h)), the ability to distinguish features between ships and complex nearshore backgrounds is limited. Due to the strong scattering interference in the nearshore area of SAR images, models without OAEM are prone to confuse the prominent features of ships with the background, resulting in blurred target boundaries or even missed detections.
After adding OAEM (Fig. \ref{fig:ssddvisfeature} (e)), the ship features in the feature map become more prominent, and the separation from the background is significantly improved. Especially in complex backgrounds, OAEM enhances the ability to extract geometric and semantic features of ships, making the expression of targets in the feature space clearer.

3. Dense distribution of ship scenes.
In the feature map without OAEM (Fig. \ref{fig:ssddvisfeature} (i)), densely distributed ship features are prone to overlap and confusion, resulting in some target features being ignored or fused together. The boundaries of ships in the feature map appear blurry, making it difficult to accurately distinguish adjacent targets.
After adding OAEM (Fig. \ref{fig:ssddvisfeature} (f)), the densely distributed ship features were significantly enhanced. The features of each vessel are more independent, the boundaries are clearer, the segmentation between adjacent vessels is significantly improved, and the detailed features of the target are better preserved.

The ablation experiment shows that OAEM significantly improves the performance of the model in SAR ship instance segmentation tasks through the organic combination of direction sensitive feature extraction, polar geometric priors, and dynamic feature fusion. Its design solves the segmentation problems of directional diversity, complex background interference, and dense scenes, providing an efficient and robust solution for SAR image ship instance segmentation.

\section{CONCLUSION}\label{sec:con}

In this study, we propose O\textsuperscript{2}Former, a new framework for ship instance segmentation in SAR images, aimed at addressing the challenges of multi-scale, dense objects, and fuzzy target boundary. In order to overcome the problems in SAR ship instance segmentation tasks, we propose OQG to integrate multi-scale features into the query vector, making the entire set of queries more sensitive to ships of various sizes. In addition, to adapt to the multi directionality of ships on the sea surface, we propose OAEM, which captures different directional features through adaptive directional convolution, enriching the features of pixel decoders. The experimental evaluation of the public datasets SSDD and HRSID confirmed the superior performance of the model, especially in detecting small objects and maintaining high localization accuracy under complex background conditions. The adaptability of this model in different scenarios highlights its practical deployment potential in maritime surveillance and related applications. Future work will focus on further optimizing computational efficiency and expanding its applicability to other remote sensing tasks.

\bibliographystyle{IEEEtran}
\bibliography{ref}

\begin{thebibliography}{10}
\providecommand{\url}[1]{#1}
\csname url@samestyle\endcsname
\providecommand{\newblock}{\relax}
\providecommand{\bibinfo}[2]{#2}
\providecommand{\BIBentrySTDinterwordspacing}{\spaceskip=0pt\relax}
\providecommand{\BIBentryALTinterwordstretchfactor}{4}
\providecommand{\BIBentryALTinterwordspacing}{\spaceskip=\fontdimen2\font plus
\BIBentryALTinterwordstretchfactor\fontdimen3\font minus
  \fontdimen4\font\relax}
\providecommand{\BIBforeignlanguage}[2]{{%
\expandafter\ifx\csname l@#1\endcsname\relax
\typeout{** WARNING: IEEEtran.bst: No hyphenation pattern has been}%
\typeout{** loaded for the language `#1'. Using the pattern for}%
\typeout{** the default language instead.}%
\else
\language=\csname l@#1\endcsname
\fi
#2}}
\providecommand{\BIBdecl}{\relax}
\BIBdecl

\bibitem{mafei}
F.~Ma, Y.~Feng, F.~Zhang, and Y.~Zhou, ``Cloud adversarial example generation
  for remote sensing image classification,'' \emph{IEEE Transactions on
  Geoscience and Remote Sensing}, 2025.

\bibitem{langrongling}
L.~Rongling, X.~Hao, G.~Fei, T.~Zewen, W.~Zhipeng \emph{et~al.}, ``Improving
  doa estimation of gnss interference through sparse non-uniform array
  reconfiguration,'' \emph{Chinese Journal of Aeronautics}, p. 103384, 2024.

\bibitem{huangheqing}
H.~Huang, F.~Gao, J.~Sun, J.~Wang, A.~Hussain, and H.~Zhou, ``Novel category
  discovery without forgetting for automatic target recognition,'' \emph{IEEE
  Journal of Selected Topics in Applied Earth Observations and Remote Sensing},
  vol.~17, pp. 4408--4420, 2024.

\bibitem{10636068}
L.~Kong, F.~Gao, X.~He, J.~Wang, J.~Sun, H.~Zhou, and A.~Hussain, ``Few-shot
  class-incremental sar target recognition via orthogonal distributed
  features,'' \emph{IEEE Transactions on Aerospace and Electronic Systems},
  vol.~61, no.~1, pp. 325--341, 2025.

\bibitem{ma2021fast}
F.~Ma, F.~Zhang, Q.~Yin, D.~Xiang, and Y.~Zhou, ``Fast sar image segmentation
  with deep task-specific superpixel sampling and soft graph convolution,''
  \emph{IEEE Transactions on Geoscience and Remote Sensing}, vol.~60, pp.
  1--16, 2021.

\bibitem{shan2023sar}
H.~Shan, X.~Fu, Z.~Lv, and Y.~Zhang, ``Sar ship detection algorithm based on
  deep dense sim attention mechanism network,'' \emph{IEEE Sensors Journal},
  vol.~23, no.~14, pp. 16\,032--16\,041, 2023.

\bibitem{zhongfengjun}
F.~Zhong, F.~Gao, T.~Liu, J.~Wang, J.~Sun, and H.~Zhou, ``Scattering
  characteristics guided network for isar space target component
  segmentation,'' \emph{IEEE Geoscience and Remote Sensing Letters}, 2025.

\bibitem{xu2017gland}
Y.~Xu, Y.~Li, Y.~Wang, M.~Liu, Y.~Fan, M.~Lai, and E.~I.-C. Chang, ``Gland
  instance segmentation using deep multichannel neural networks,'' \emph{IEEE
  Transactions on Biomedical Engineering}, vol.~64, no.~12, pp. 2901--2912,
  2017.

\bibitem{zheng2021enhancing}
Z.~Zheng, P.~Wang, D.~Ren, W.~Liu, R.~Ye, Q.~Hu, and W.~Zuo, ``Enhancing
  geometric factors in model learning and inference for object detection and
  instance segmentation,'' \emph{IEEE transactions on cybernetics}, vol.~52,
  no.~8, pp. 8574--8586, 2021.

\bibitem{zhaowei}
X.~Shang, N.~Li, D.~Li, J.~Lv, W.~Zhao, R.~Zhang, and J.~Xu, ``Ccldet: A
  cross-modality and cross-domain low-light detector,'' \emph{IEEE Transactions
  on Intelligent Transportation Systems}, 2025.

\bibitem{xu2023edge}
S.~Xu, J.~Fan, X.~Jia, and J.~Chang, ``Edge-constrained guided feature
  perception network for ship detection in sar images,'' \emph{IEEE Sensors
  Journal}, vol.~23, no.~21, pp. 26\,828--26\,838, 2023.

\bibitem{li2024multiscale}
Z.~Li, D.~Kong, J.~Liu, X.~Sun, Q.~Du, and L.~Zhang, ``Multiscale accurate ship
  detection network driven by multiattention fusion for complex maritime
  backgrounds,'' \emph{IEEE Sensors Journal}, vol.~24, no.~6, pp. 9208--9216,
  2024.

\bibitem{li2017improved}
T.~Li, Z.~Liu, R.~Xie, and L.~Ran, ``An improved superpixel-level cfar
  detection method for ship targets in high-resolution sar images,'' \emph{IEEE
  Journal of Selected Topics in Applied Earth Observations and Remote Sensing},
  vol.~11, no.~1, pp. 184--194, 2017.

\bibitem{Mask2Former}
B.~Cheng, I.~Misra, A.~G. Schwing, A.~Kirillov, and R.~Girdhar,
  ``Masked-attention mask transformer for universal image segmentation,'' in
  \emph{Proceedings of the IEEE/CVF conference on computer vision and pattern
  recognition}, 2022, pp. 1290--1299.

\bibitem{threshold1}
L.~D. Pacifico, T.~B. Ludermir, and L.~F. Britto, ``A hybrid improved group
  search optimization and otsu method for color image segmentation,'' in
  \emph{2018 7th Brazilian Conference on Intelligent Systems (BRACIS)}.\hskip
  1em plus 0.5em minus 0.4em\relax IEEE, 2018, pp. 296--301.

\bibitem{edge1}
R.~Fjortoft, A.~Lopes, P.~Marthon, and E.~Cubero-Castan, ``An optimal multiedge
  detector for sar image segmentation,'' \emph{IEEE Transactions on Geoscience
  and Remote Sensing}, vol.~36, no.~3, pp. 793--802, 1998.

\bibitem{region1}
S.~Fan, Y.~Sun, and P.~Shui, ``Region-merging method with texture pattern
  attention for sar image segmentation,'' \emph{IEEE Geoscience and Remote
  Sensing Letters}, vol.~18, no.~1, pp. 112--116, 2020.

\bibitem{region2}
F.~Ma, F.~Gao, J.~Sun, H.~Zhou, and A.~Hussain, ``Weakly supervised
  segmentation of sar imagery using superpixel and hierarchically adversarial
  crf,'' \emph{Remote Sensing}, vol.~11, no.~5, p. 512, 2019.

\bibitem{SSDD}
T.~Zhang, X.~Zhang, J.~Li, X.~Xu, B.~Wang, X.~Zhan, Y.~Xu, X.~Ke, T.~Zeng,
  H.~Su \emph{et~al.}, ``Sar ship detection dataset (ssdd): Official release
  and comprehensive data analysis,'' \emph{Remote Sensing}, vol.~13, no.~18, p.
  3690, 2021.

\bibitem{HRSID}
S.~Wei, X.~Zeng, Q.~Qu, M.~Wang, H.~Su, and J.~Shi, ``Hrsid: A high-resolution
  sar images dataset for ship detection and instance segmentation,'' \emph{Ieee
  Access}, vol.~8, pp. 120\,234--120\,254, 2020.

\bibitem{HTC}
K.~Chen, J.~Pang, J.~Wang, Y.~Xiong, X.~Li, S.~Sun, W.~Feng, Z.~Liu, J.~Shi,
  W.~Ouyang \emph{et~al.}, ``Hybrid task cascade for instance segmentation,''
  in \emph{Proceedings of the IEEE/CVF conference on computer vision and
  pattern recognition}, 2019, pp. 4974--4983.

\bibitem{PANet}
S.~Liu, L.~Qi, H.~Qin, J.~Shi, and J.~Jia, ``Path aggregation network for
  instance segmentation,'' in \emph{Proceedings of the IEEE conference on
  computer vision and pattern recognition}, 2018, pp. 8759--8768.

\bibitem{he2017mask}
K.~He, G.~Gkioxari, P.~Doll{\'a}r, and R.~Girshick, ``Mask r-cnn,'' in
  \emph{Proceedings of the IEEE international conference on computer vision},
  2017, pp. 2961--2969.

\bibitem{lin2017feature}
T.-Y. Lin, P.~Doll{\'a}r, R.~Girshick, K.~He, B.~Hariharan, and S.~Belongie,
  ``Feature pyramid networks for object detection,'' in \emph{Proceedings of
  the IEEE conference on computer vision and pattern recognition}, 2017, pp.
  2117--2125.

\bibitem{CascadeRCNN}
Z.~Cai and N.~Vasconcelos, ``Cascade r-cnn: Delving into high quality object
  detection,'' in \emph{Proceedings of the IEEE conference on computer vision
  and pattern recognition}, 2018, pp. 6154--6162.

\bibitem{MS-RCNN}
Z.~Huang, L.~Huang, Y.~Gong, C.~Huang, and X.~Wang, ``Mask scoring r-cnn,'' in
  \emph{Proceedings of the IEEE/CVF conference on computer vision and pattern
  recognition}, 2019, pp. 6409--6418.

\bibitem{Solo}
X.~Wang, T.~Kong, C.~Shen, Y.~Jiang, and L.~Li, ``Solo: Segmenting objects by
  locations,'' in \emph{Computer Vision--ECCV 2020: 16th European Conference,
  Glasgow, UK, August 23--28, 2020, Proceedings, Part XVIII 16}.\hskip 1em plus
  0.5em minus 0.4em\relax Springer, 2020, pp. 649--665.

\bibitem{Yolact}
D.~Bolya, C.~Zhou, F.~Xiao, and Y.~J. Lee, ``Yolact: Real-time instance
  segmentation,'' in \emph{Proceedings of the IEEE/CVF international conference
  on computer vision}, 2019, pp. 9157--9166.

\bibitem{HQISNET}
H.~Su, S.~Wei, S.~Liu, J.~Liang, C.~Wang, J.~Shi, and X.~Zhang, ``Hq-isnet:
  High-quality instance segmentation for remote sensing imagery,'' \emph{Remote
  Sensing}, vol.~12, no.~6, p. 989, 2020.

\bibitem{SISNet}
Z.~Shao, X.~Zhang, S.~Wei, J.~Shi, X.~Ke, X.~Xu, X.~Zhan, T.~Zhang, and
  T.~Zeng, ``Scale in scale for sar ship instance segmentation,'' \emph{Remote
  Sensing}, vol.~15, no.~3, p. 629, 2023.

\bibitem{CATNet}
Y.~Liu, H.~Li, C.~Hu, S.~Luo, Y.~Luo, and C.~W. Chen, ``Learning to aggregate
  multi-scale context for instance segmentation in remote sensing images,''
  \emph{IEEE Transactions on Neural Networks and Learning Systems}, vol.~36,
  no.~1, pp. 595--609, 2024.

\bibitem{kirillov2023segment}
A.~Kirillov, E.~Mintun, N.~Ravi, H.~Mao, C.~Rolland, L.~Gustafson, T.~Xiao,
  S.~Whitehead, A.~C. Berg, W.-Y. Lo \emph{et~al.}, ``Segment anything,'' in
  \emph{Proceedings of the IEEE/CVF international conference on computer
  vision}, 2023, pp. 4015--4026.

\bibitem{dosovitskiy2020image}
A.~Dosovitskiy, L.~Beyer, A.~Kolesnikov, D.~Weissenborn, X.~Zhai,
  T.~Unterthiner, M.~Dehghani, M.~Minderer, G.~Heigold, S.~Gelly \emph{et~al.},
  ``An image is worth 16x16 words: Transformers for image recognition at
  scale,'' \emph{arXiv preprint arXiv:2010.11929}, 2020.

\bibitem{DETR}
N.~Carion, F.~Massa, G.~Synnaeve, N.~Usunier, A.~Kirillov, and S.~Zagoruyko,
  ``End-to-end object detection with transformers,'' in \emph{European
  conference on computer vision}.\hskip 1em plus 0.5em minus 0.4em\relax
  Springer, 2020, pp. 213--229.

\bibitem{maskformer}
B.~Cheng, I.~Misra, A.~G. Schwing, A.~Kirillov, and R.~Girdhar,
  ``Masked-attention mask transformer for universal image segmentation,'' in
  \emph{Proceedings of the IEEE/CVF conference on computer vision and pattern
  recognition}, 2022, pp. 1290--1299.

\bibitem{Segmenter}
R.~Strudel, R.~Garcia, I.~Laptev, and C.~Schmid, ``Segmenter: Transformer for
  semantic segmentation,'' in \emph{Proceedings of the IEEE/CVF international
  conference on computer vision}, 2021, pp. 7262--7272.

\bibitem{Segformer}
E.~Xie, W.~Wang, Z.~Yu, A.~Anandkumar, J.~M. Alvarez, and P.~Luo, ``Segformer:
  Simple and efficient design for semantic segmentation with transformers,''
  \emph{Advances in neural information processing systems}, vol.~34, pp.
  12\,077--12\,090, 2021.

\bibitem{Rsprompter}
K.~Chen, C.~Liu, H.~Chen, H.~Zhang, W.~Li, Z.~Zou, and Z.~Shi, ``Rsprompter:
  Learning to prompt for remote sensing instance segmentation based on visual
  foundation model,'' \emph{IEEE Transactions on Geoscience and Remote
  Sensing}, vol.~62, pp. 1--17, 2024.

\bibitem{EISP}
F.~Fan, X.~Zeng, S.~Wei, H.~Zhang, D.~Tang, J.~Shi, and X.~Zhang, ``Efficient
  instance segmentation paradigm for interpreting sar and optical images,''
  \emph{Remote Sensing}, vol.~14, no.~3, p. 531, 2022.

\bibitem{shen2025imaggarment}
F.~Shen, J.~Yu, C.~Wang, X.~Jiang, X.~Du, and J.~Tang, ``Imaggarment-1:
  Fine-grained garment generation for controllable fashion design,''
  \emph{arXiv preprint arXiv:2504.13176}, 2025.

\bibitem{dabdetr}
S.~Liu, F.~Li, H.~Zhang, X.~Yang, X.~Qi, H.~Su, J.~Zhu, and L.~Zhang,
  ``Dab-detr: Dynamic anchor boxes are better queries for detr,'' \emph{arXiv
  preprint arXiv:2201.12329}, 2022.

\bibitem{dino}
H.~Zhang, F.~Li, S.~Liu, L.~Zhang, H.~Su, J.~Zhu, L.~M. Ni, and H.-Y. Shum,
  ``Dino: Detr with improved denoising anchor boxes for end-to-end object
  detection,'' \emph{arXiv preprint arXiv:2203.03605}, 2022.

\bibitem{fastinst}
J.~He, P.~Li, Y.~Geng, and X.~Xie, ``Fastinst: A simple query-based model for
  real-time instance segmentation,'' in \emph{Proceedings of the IEEE/CVF
  conference on computer vision and pattern recognition}, 2023, pp.
  23\,663--23\,672.

\bibitem{shen2025imagdressing}
F.~Shen, X.~Jiang, X.~He, H.~Ye, C.~Wang, X.~Du, Z.~Li, and J.~Tang,
  ``Imagdressing-v1: Customizable virtual dressing,'' in \emph{Proceedings of
  the AAAI Conference on Artificial Intelligence}, vol.~39, no.~7, 2025, pp.
  6795--6804.

\bibitem{shen2024imagpose}
F.~Shen and J.~Tang, ``Imagpose: A unified conditional framework for
  pose-guided person generation,'' \emph{Advances in neural information
  processing systems}, vol.~37, pp. 6246--6266, 2024.

\bibitem{shen2025long}
F.~Shen, C.~Wang, J.~Gao, Q.~Guo, J.~Dang, J.~Tang, and T.-S. Chua, ``Long-term
  talkingface generation via motion-prior conditional diffusion model,''
  \emph{arXiv preprint arXiv:2502.09533}, 2025.

\bibitem{jiang2023dawn}
K.~Jiang, W.~Liu, Z.~Wang, X.~Zhong, J.~Jiang, and C.-W. Lin, ``Dawn:
  Direction-aware attention wavelet network for image deraining,'' in
  \emph{Proceedings of the 31st ACM international conference on multimedia},
  2023, pp. 7065--7074.

\bibitem{kong2023direction}
W.~Kong, J.~Shen, H.~Li, J.~Liu, and J.~Zhang, ``Direction-aware attention
  aggregation for single-stage hazy-weather crowd counting,'' \emph{Expert
  Systems with Applications}, vol. 225, p. 120088, 2023.

\bibitem{chen2021building}
K.~Chen, Z.~Zou, and Z.~Shi, ``Building extraction from remote sensing images
  with sparse token transformers,'' \emph{Remote Sensing}, vol.~13, no.~21, p.
  4441, 2021.

\bibitem{cheng2024direction}
H.~Cheng, Y.~Cao, B.~Sui, S.~Zhang, and Y.~Zeng, ``Direction-aware multi-branch
  attention and gaussian label assignment for remote sensing aggregative object
  detection,'' \emph{International Journal of Remote Sensing}, vol.~45, no.~17,
  pp. 5917--5945, 2024.

\bibitem{chen2023arbitrary}
Y.~Chen, J.~Wang, Y.~Zhang, and Y.~Liu, ``Arbitrary-oriented ship detection
  based on kullback-leibler divergence regression in remote sensing images,''
  \emph{Earth Science Informatics}, vol.~16, no.~4, pp. 3243--3255, 2023.

\bibitem{shen2023advancing}
F.~Shen, H.~Ye, J.~Zhang, C.~Wang, X.~Han, and W.~Yang, ``Advancing pose-guided
  image synthesis with progressive conditional diffusion models,'' \emph{arXiv
  preprint arXiv:2310.06313}, 2023.

\bibitem{shen2025boosting}
F.~Shen, H.~Ye, S.~Liu, J.~Zhang, C.~Wang, X.~Han, and Y.~Wei, ``Boosting
  consistency in story visualization with rich-contextual conditional diffusion
  models,'' in \emph{Proceedings of the AAAI Conference on Artificial
  Intelligence}, vol.~39, no.~7, 2025, pp. 6785--6794.

\bibitem{coco}
T.-Y. Lin, M.~Maire, S.~Belongie, J.~Hays, P.~Perona, D.~Ramanan,
  P.~Doll{\'a}r, and C.~L. Zitnick, ``Microsoft coco: Common objects in
  context,'' in \emph{Computer vision--ECCV 2014: 13th European conference,
  zurich, Switzerland, September 6-12, 2014, proceedings, part v 13}.\hskip 1em
  plus 0.5em minus 0.4em\relax Springer, 2014, pp. 740--755.

\bibitem{tian2020condistl}
Z.~Tian, C.~Shen, and H.~Chen, ``Conditional convolutions for instance
  segmentation,'' in \emph{Computer Vision--ECCV 2020: 16th European
  Conference, Glasgow, UK, August 23--28, 2020, Proceedings, Part I 16}.\hskip
  1em plus 0.5em minus 0.4em\relax Springer, 2020, pp. 282--298.

\bibitem{rtmdet}
C.~Lyu, W.~Zhang, H.~Huang, Y.~Zhou, Y.~Wang, Y.~Liu, S.~Zhang, and K.~Chen,
  ``Rtmdet: An empirical study of designing real-time object detectors,''
  \emph{arXiv preprint arXiv:2212.07784}, 2022.

\bibitem{AFSS}
F.~Gao, Y.~Huo, J.~Wang, A.~Hussain, and H.~Zhou, ``Anchor-free sar ship
  instance segmentation with centroid-distance based loss,'' \emph{IEEE Journal
  of Selected Topics in Applied Earth Observations and Remote Sensing},
  vol.~14, pp. 11\,352--11\,371, 2021.

\bibitem{srnet}
X.~Yang, Q.~Zhang, Q.~Dong, Z.~Han, X.~Luo, and D.~Wei, ``Ship instance
  segmentation based on rotated bounding boxes for sar images,'' \emph{Remote
  Sensing}, vol.~15, no.~5, p. 1324, 2023.

\bibitem{diffsarshipinst}
X.~Xu, X.~Zhang, S.~Wei, J.~Shi, W.~Zhang, T.~Zhang, X.~Zhan, Y.~Xu, and
  T.~Zeng, ``Diffsarshipinst: Diffusion model for ship instance segmentation
  from synthetic aperture radar imagery,'' \emph{ISPRS Journal of
  Photogrammetry and Remote Sensing}, vol. 223, pp. 440--455, 2025.

\end{thebibliography}

\begin{IEEEbiography}[{\includegraphics[width=1in,height=1.25in,clip,keepaspectratio]{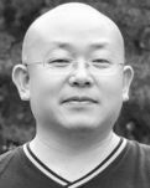}}]{Fei Gao} received the B.S. degree in industrial electrical automation and the M.S. degree in electromagnetic measurement technology and instrument from Xi'an Petroleum Institute, Xi'an, China, in 1996 and 1999, respectively, and the Ph.D. degree in signal and information processing from Beihang University, Beijing, China, in 2005.

He is currently a Professor with the School of Electronic and Information Engineering, Beihang University. His research interests include target detection and recognition, image processing, and deep learning for applications in remote sensing.
\end{IEEEbiography}

\begin{IEEEbiography}[{\includegraphics[width=1in,height=1.25in,clip,keepaspectratio]{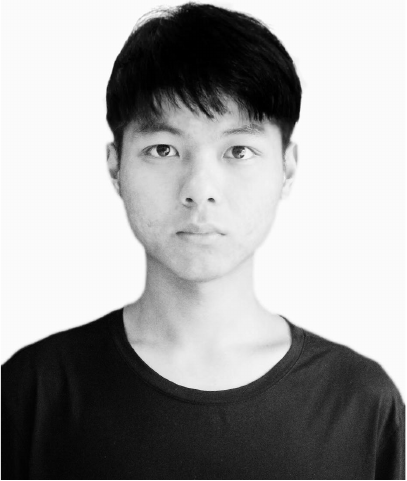}}]{Yunrui Li} received the B.S. degree in communication engineering from Jilin University, Changchun, China, in 2023. He is currently pursuing the M.S. degree in electronic information engineering at Beihang University. His current research activities are in target instance segmentation and synthetic aperture radar image processing.
\end{IEEEbiography}

\begin{IEEEbiography}[{\includegraphics[width=1in,height=1.25in,clip,keepaspectratio]{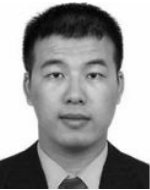}}]{Xiaoyu He} was born in Jiangxi, China, in 1991. He received the B.S. degree in information engineering from the School of Mechanical and Electric Engineering,Guangzhou University, Guangzhou, China, in 2012, and the Ph.D. degree in signal and information processing from the School of Electronics and Information Engineering, Beihang University. Beijing, China, in 2018.

He is currently a Lecturer with the School of Electronics and Information Engineering, Beihang University, His research interests include characteristic modeling for multi spectral and hyperspectral imagery, radar system modeling, and signal processing.
\end{IEEEbiography}

\begin{IEEEbiography}[{\includegraphics[width=1in,height=1.25in,clip,keepaspectratio]{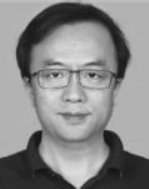}}]{Jinping Sun} received the M.Sc.degree in communication and information systems and Ph.D. degree in information and communication engineering from Beihang University, Beijing, China, in 1998 and 2001, respectively.

He is currently a Professor with the School of Electronic and Information Engineering, Beihang University. His research interests include statistical signal processing, high-resolution radar signal processing,target tracking, image understanding, and robust beamforming.
\end{IEEEbiography}

\begin{IEEEbiography}[{\includegraphics[width=1in,height=1.25in,clip,keepaspectratio]{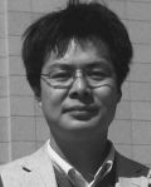}}]{Jun Wang} received the B.S. degree in communication engineering from North Western Polytechnical University, Xi'an, China, in 1995, and the M.S. and Ph.D,degrees in signal and information processing from the Beijing University of Aeronautics and Astronautics (now Beihang University), Beijing, China, in 1998 and 2001, respectively.

He is currently a Professor with the School of Electronic and Information Engineering, Beihang University. He authored or coauthored more than 40 papers in journals, books, and conference proceedings. His research interests include signal processing, digital signal real-time architecture, and target processor/field programmable gate array recognition and tracking.
\end{IEEEbiography}

\end{document}